\documentclass[fleqn,10pt]{wlscirep}
\usepackage[utf8]{inputenc}
\usepackage[T1]{fontenc}
\usepackage{amsmath}
\usepackage{float}
\usepackage{multirow}
\usepackage[linesnumbered,ruled]{algorithm2e}
\usepackage{graphicx}  
\usepackage{float}  
\usepackage{subfigure}  

\title{Knowledge Relation Rank Enhanced Heterogeneous Learning Interaction Modeling for Neural Graph Forgetting Knowledge Tracing}

\author[1,+]{Linqing Li}
\author[1,2,+,*]{Zhifeng Wang}
\affil[1]{Central China Normal University Wollongong Joint Institute, Central China Normal University, Wuhan 430079, China}
\affil[2]{Faculty of Artificial Intelligence in Education, Central China Normal University, Wuhan 430079, China}

\affil[*]{zfwang@ccnu.edu.cn}

\affil[+]{these authors contributed equally to this work}

\keywords{Knowledge Tracing, Knowledge Relation Rank, Heterogeneous Graph, Forgetting Behavior, Attention Mechanism}

\begin{abstract}
Recently, knowledge tracing models have been applied in educational data mining such as the Self-attention knowledge tracing model(SAKT), which models the relationship between exercises and Knowledge concepts(Kcs). However, relation modeling in traditional Knowledge tracing models only considers the static question-knowledge relationship and knowledge-knowledge relationship and treats these relationships with equal importance. This kind of relation modeling is difficult to avoid the influence of subjective labeling and considers the relationship between exercises and KCs, or KCs and KCs separately.
In this work, a novel knowledge tracing model, named Knowledge Relation Rank Enhanced Heterogeneous Learning Interaction Modeling for Neural Graph Forgetting Knowledge Tracing(NGFKT), is proposed to reduce the impact of the subjective labeling by calibrating the skill relation matrix and the Q-matrix and apply the Graph Convolutional Network(GCN) to model the heterogeneous interactions between students, exercises, and skills. Specifically, the skill relation matrix and Q-matrix are generated by the Knowledge Relation Importance Rank Calibration method(KRIRC). Then the calibrated skill relation matrix, Q-matrix, and the heterogeneous interactions are treated as the input of the GCN to generate the exercise embedding and skill embedding. Next, the exercise embedding, skill embedding, item difficulty, and contingency table are incorporated to generate an exercise relation matrix as the inputs of the Position-Relation-Forgetting attention mechanism. Finally, the Position-Relation-Forgetting attention mechanism is applied to make the predictions. Experiments are conducted on the two public educational datasets and results indicate that the NGFKT model outperforms all baseline models in terms of AUC, ACC, and Performance Stability(PS).
\end{abstract}
\begin{document}

\flushbottom
\maketitle

\thispagestyle{empty}
\section*{Introduction}
With the continuous development of science and technology, the network provides people with many conveniences and created huge amounts of information. Our modern generation is defined by the rapid growth of big data, which is crucial to connecting many areas, including education, health care, and transportation. Combining education with information science is an unstoppable trend in the development of education topics, and the use of various technologies developing in online education has expanded into a significant field of research. The application of artificial intelligence and related technologies to analyze large amounts of educational data generates valuable information for people from it and processes it to further education. Online educational systems have been widely applied in the educational field for tracking, reporting, and the delivery of online courses including edX, Coursera, and Udacity\cite{hamid2021content}. For students, these platforms provide a variety of conveniences including several online courses and free, individualized learning resources \cite{n2020distance}. Furthermore, these online learning platforms support the normal teaching schedule because online teaching is difficult to carry out in many regions due to geographical reasons or weather reasons. These powerful educational systems can be utilized by teachers to create remedial materials based on the needs of their students\cite{de2020twelve}. Tracking the students' performance based on past interactions is proved as an important task\cite{tomasevic2020overview} in these educational systems. This task is known as the \emph{Knowledge tracing} \cite{2021A} which aims to analyze exercises to infer the knowledge state of the students by their responses. \emph{Knowledge tracing}(KT) can be considered as the task to evaluate the performance of the student knowledge state. Specifically, students can first select a set of questions $X_t$ = $(x_1, x_2,...x_t)$ to practice the knowledge points and response logs (e.g., right or wrong). Then, the objective of the KT is the probability of correctly predicting the knowledge states of the students in the next interaction $p(a_{t+1}=1|q_{t+1}, X)$ according to past interactions and corresponding responses. The input: x$_t$ is presented as the tuple($q_t$,a$_t$), where $q_t$ infers the question of the student and a$_t$ indicates the response provided by the student.

The Q-matrix is served as the process to model the learning resources used in many models\cite{huang2020learning}. The value of the Q-matrix is binary indicating the relationship between exercises and KCs. Specifically, the value of the matrix is marked as 1 when the problem is related to the KC, otherwise, it is marked as 0. However, the relationship between skills is ignored. Therefore, there exist several calibration methods to generate the calibrated Q-matrix but those methods ignore the knowledge importance between Knowledge Concepts(KCs)\cite{wang2022tracking}.

Recently, numerous knowledge tracing models are designed to handle the problem of tracking the knowledge state of students, such as the Deep Knowledge Tracing model(DKT)\cite{piech2015deep}, the Dynamic Key-Value Memory network(DKVMN)\cite{zhang2017dynamic}, Graph-based knowledge tracing(GKT)\cite{nakagawa2019graph}, and the Self-attentive model for knowledge tracing(SAKT) \cite{pandey2019self}. These models achieve better prediction performance in several public educational datasets. However, the DKT model ignores the information about the knowledge points and can not take the students' abilities into consideration. The SAKT and DKVMN further consider the relationship between exercises and skills without considering the heterogeneous interactions of the students, exercises, and KCs. The GKT is proposed to consider the heterogeneous interactions of students, exercises, and KCs but ignores the relational information between exercises and KCs. 

In order to solve the problem of previous models, the Knowledge Relation Rank Enhanced Heterogeneous Learning Interaction Modeling for Neural Graph Forgetting Knowledge Tracing(NGFKT) is designed to address the drawbacks of these four models. Firstly, the skill relation matrix and Q-matrix are generated and calibrated to consider the relationship between questions and KCs. Then the calibrated skill relation matrix and the Q-matrix are served as the inputs of the GCN to output the skill-exercise embedding. Next, the heterogeneous interactions, the item difficulty, the skill embedding, and the exercise embedding are incorporated to model the exercise relation matrix. In addition, the Position-Relation-Forgetting attention mechanism is applied to predict the students' performance based on the exercise relation matrix.

The main contributions of this paper are:
\begin{enumerate}
    \item A calibration method is proposed to calibrate the skill relation matrix and Q-matrix according to the relationship of KCs in the heterogeneous interactions:students-exercise-KCs. Then, the calibrated skill relation matrix, Q-matrix, and the heterogeneous interactions are used for the input of the GCN to generate the exercise embedding and skill embedding, which comprehensively considers the relationship between students, exercises, and KCs.
    \item A exercise relation modeling method is designed to obtain the exercise relation matrix by incorporating the exercise embedding, skill embedding, and item difficulty with the contingency table. Compared with the GKT model \cite{nakagawa2019graph}, our proposed model considers the student's forgetting behavior and relative distance representations by applying the Position-Relation-Forgetting attention mechanism to make predictions. The implementation codes of this research are available at \url{https://github.com/Destiny123456qwer/NGFKT}.
    \item Detailed experiments are conducted to evaluate the performance between the NGFKT model and baseline models from three aspects. The first is to measure the prediction performance of the NGFKT model with the baseline model and a new metric: PS is designed  to measure the performance with AUC and ACC. The second is to find out the effectiveness of the NGFKT model even with limited records. The last aspect is to visualize the knowledge tracing results based on the radar diagrams.
\end{enumerate}
The rest of the paper is organized as follows. The related works on the knowledge tracing models, graph neural networks, and relation modeling are included in the "Related works". The section "Proposed model" describes the structure of the NGFKT model in detail. In the "Implementation and experimental results" section, the experimental results of this paper and implementation details are presented. Finally, the conclusion and future work directions are shown in the section"Conclusion and future work"
\section*{Related works}
\paragraph{Knowledge tracing.} The knowledge tracking task seeks to assess the students' level of knowledge according to the students' interactions. Numerous KT models created on deep learning now lead to improvements in tracking students' knowledge states. The Deep Knowledge Tracing model(DKT)\cite{piech2015deep} is the first method to employ a neural network to track the knowledge state of the students. The effectiveness of the DKT model also is further verified \cite{xiong2016going}. However, most knowledge-tracing models based on the deep neural network ignore the relationship between exercises and skills. The EKT model is proposed to apply the exercise embedding modules to model the relation between the exercises\cite{liu2019ekt}. However, the EKT model does not take past interactions and student behavior into consideration. The RKT model combines past interactions with the attention mechanism to make predictions\cite{pandey2020rkt} without considering the relationship between students, exercises, and KCs.

\paragraph{Graph neural networks.} A non-linear conducted graph is more complicated than a tree structure. Information with a graph structure can describe connections and entities in the real world in an intuitive manner\cite{battaglia2018relational}. A unique type of neural network called a GCN operates directly on data with a graph structure. The GCN model can update the presentation of the nodes in the graph by itself and their neighbors\cite{kipf2016semi}. Applying numerous graph convolutional layers, the GCN model can make sure that the updated nodes properly represent both the features of higher-order neighbors and the features of neighboring nodes.

 There exist some knowledge tracing models focusing on investigating the potential usage of knowledge graph structures to capture more knowledge. According to the works of \cite{chanaa2020predicting}, each node represents a student; the nodes are fixed, but the edges are dynamic. This creates a dynamic graph that changes over time. However, this method ignores the heterogeneous interactions between the students-exercises-KCs. The GIKT model is proposed to treat the exercise-skill relationship graph as a bipartite graph and then applies the embedding propagation to the GCN to integrate the exercise-skill correlation\cite{yang2021gikt} without considering student behavior. Therefore, the NGFKT model applies the skill relation matrix and Q-matrix to model the relationship between exercises and KCs and predicts the performance by incorporating the attention mechanism with a forgetting layer.
\paragraph{Relation modeling.} The relation modeling is also involved in the knowledge tracing tasks including the exercise relation modeling. Exercise relation modeling is applied in many research papers to make predictions\cite{song2020sepn}. When two exercises are associated with the same knowledge concept or the student, they can be considered connected\cite{wu2022sgkt}. In addition, the Q-matrix is used for investigating the relationship between exercises and skills to explore the connection between two exercises\cite{nagatani2019augmenting}. Intuitively, if the two problems are similar in difficulty and similar in past practice, then the two problems can also be considered similar\cite{gan2022knowledge,pandey2020rkt}. According to these ideas, the NGFKT model incorporates the item difficulty, the exercise embedding, the skill embedding, and the contingency table of two exercises to generate the final exercise relation matrix.

\section*{Proposed model\label{pm}}
In this section, the NGFKT model is developed based on the exercise relation matrix and Position-Relation-Forgetting attention mechanism to predict the performance of students. There exist three steps including the skill relation matrix and Q-matrix modeling, the extraction of the exercise relation matrix, and the predictions based on the Position-Relation-Forgetting attention mechanism. First, the skill relation matrix($\hat{S}$) and the calibrated Q-matrix ($\hat{Q}$) are designed as the input of the model. Then, the skill relation matrix $\hat{S}$, the calibrated Q-matrix ($\hat{Q}$), and the heterogeneous interactions are treated as the inputs of the GCN to output the embedding of exercises and skills. The similarity of exercises can be computed based on the exercise embedding, skill embedding, item difficulty, and contingency table to generate the corresponding exercise relation matrix($R^E$). Finally, the exercise relation matrix($R^E$) is served as the input of the Position-relation-Forgetting attention mechanism to track the student knowledge state. The overall structure can reference as follows:
\begin{figure}[ht]
\centering
\includegraphics[width=\linewidth]{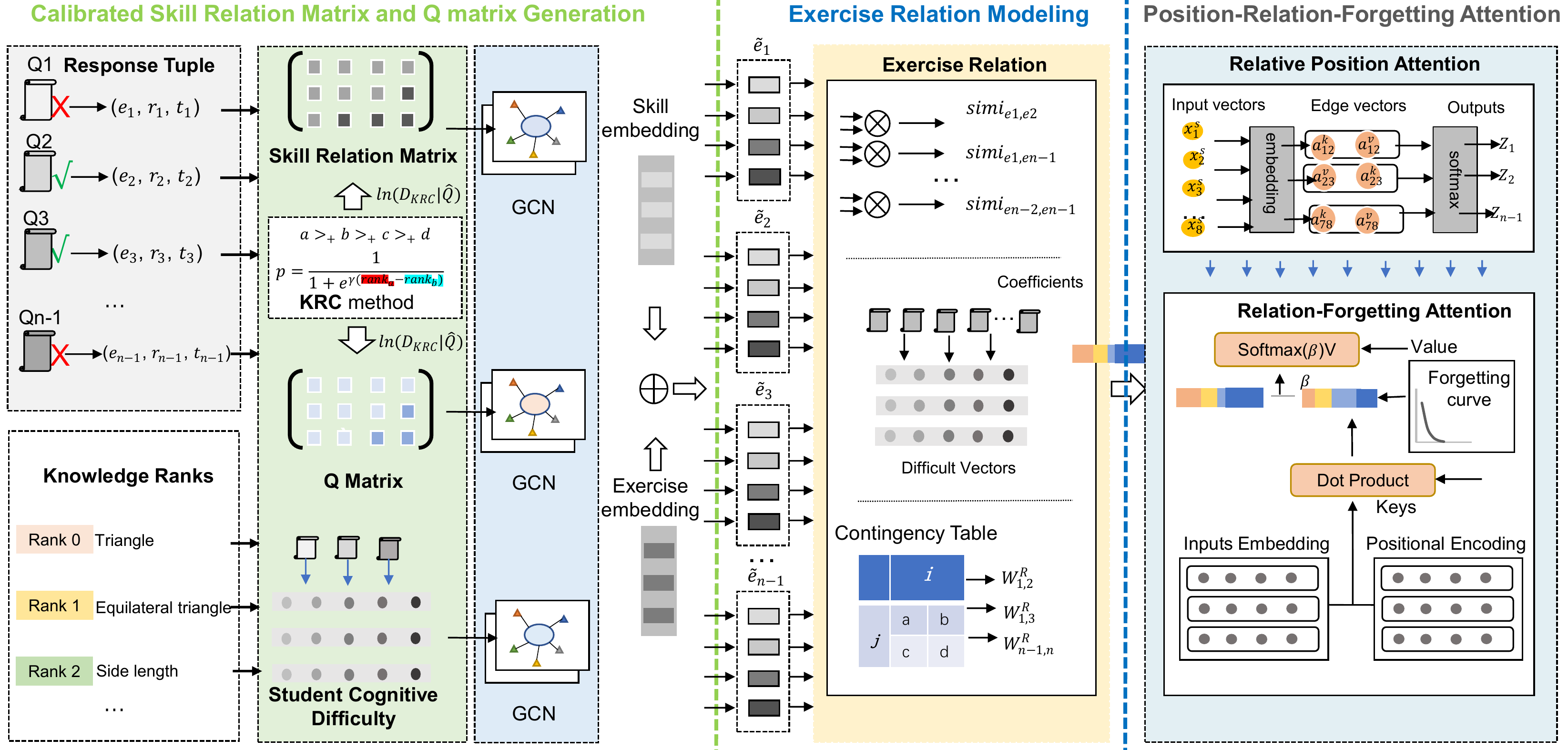}
\caption{The overall structure of the Neural Graph Forgetting Knowledge Tracing (NGFKT). Firstly, according to previous interactions, the response tuples of students and knowledge levels can be extracted from the inputs. Then KRIRC method is designed to generate the calibrated skill relation matrix and Q-matrix. The skill relation matrix, Q-matrix, and the cognitive difficulty of each student are treated as the inputs of the GCN to generate the skill-exercise embedding: $\hat{e}$. Secondly, the $\hat{e}$, the item difficulty of each question, and the contingency table are combined to output the exercise relation matrix.  Finally, the Position-Relation-Forgetting attention mechanism is utilized to process the inputs to make predictions.}
\label{fig:stream}
\end{figure}
\paragraph{Skill relation matrix and Q matrix modeling.}
The related skill can be defined as the skill that is covered by the same exercise. When considering the hierarchical knowledge levels of skill, the related skill also can be regarded as the parent nodes and child nodes of the skill. For instance, the knowledge concept: the "Triangle" is viewed as the "Right Triangle"'s parent node, and the "Pythagorean Theorem" is the child node of the "Right Triangle". Therefore, the related skills of the "Right Angle" are the concepts: "Triangle" and "Pythagorean Theorem". Inspired by these two ideas, the importance of the skills is ranked based on the following partial order, and the calibrated method called the knowledge Relation Importance Rank Calibration method is proposed(KRIRC). A pairwise Bayesian treatment is as follows. For convenience, we define a partial order $>_i^+$ as:
\begin{equation}
    a>_i^+b>_i^+>c>_i^+d  \label{p_order}
\end{equation}
where "a", "b", "c", and "d" are the neighbors of the skill. Here, "a" implies the skill of the parent node, knowledge level is 0, in the knowledge level graph. "b" denotes the skill of the child node, knowledge level is 1, in the knowledge level graph. "c" can be interpreted in a similar way. "d" is the skill that is covered by the same exercise. Along this line, neighbor: "a" is more important than neighbor: "b" in extracting the neighbor of the skill. The rank of the skill: "a", "b", "c", and "d" are 0, 1, 2, and 3 respectively. Thereby, according to the equation \ref{p_order}, the partial order relationship set can be defined as $D_{KRIRC} = \{(i, a, b)| a > _i^+ b, i = 1,2,3...K\}$ where K is the number of knowledge concepts.
 Based on the traditional Bayesian method, we assume the calibration matrix: $\hat{M}$ uniforms the Gaussian distribution. To give the calibration matrix labels higher confidence, we define $p(a >_i^+ b|\hat{M})$ as follows:
\begin{equation}
     p(a>_i^+ b|\hat{M}) = \frac{1}{1+e^{\lambda(a_{rank}-b_{rank})}} \label{c_e}
 \end{equation}
where $\lambda$ controls the discrimination of relevance values of different knowledge levels. The log posterior over D$_{KRIRC}$ on $\hat{M}$ can be eventually computed as:
\begin{eqnarray}    \label{eq}
lnp(\hat{M}|D_{KRIRC} )&=& ln\prod_{(i,a,b) \in D_{KRIRC}}^{1}p(a>_i^+b|\hat{M_i}) p(\hat{M_i}) \nonumber    \\
~&=& \sum_{i=1}^{N}\sum_{a=1}^{K}\sum_{b=1}^{K}I(a>_i^+b)ln \frac{1}{1+e^{\lambda (a_{rank}-b_{rank})}}+C-\sum_{i=1}^{N}\sum_{j=1}^{K}\frac{\hat{M}_{ij}^2 }{2\sigma ^2 } \nonumber    \\
\end{eqnarray}
where I(*) is the judgment function when the function's condition is met and the function output 1. And C, the constant variable, can be ignored when training the matrix. Finally, a calibrated matrix $\hat{M}$ estimated by the KRIRC can be calculated. The Q-matrix and the skill relation matrix can apply the KRIRC method to obtain the calibrated Q-matrix  and the calibrated skill relation to reflect the hierarchical knowledge levels between different knowledge points. The specific algorithm can reference as follows:
\begin{algorithm}[ht]
\DontPrintSemicolon
  \SetAlgoLined
  \KwIn {Students' historical response dataset: $D = {s_1, s_2,...s_N}$, $s_i=(e_i, s_i, t_i)$;
    The knowledge level graph G; 
    The heterogeneous relation graph: $\tau$;
    Task-learning rate: $\alpha$; 
    Hyper-parameter $\lambda$;}
  \KwOut {The calibrated Q matrix: $\hat{Q}$; The calibrated skill relation matrix : $\hat{S}$.}
  initialization learning rate $\alpha$ and hyper-parameter $\lambda$ randomly \;
  
  \While{element in G and $\tau$}{
    Extract hierarchical knowledge levels and related skills of each element based on G and $\tau$ \;
    Generate the corresponding knowledge rank using equation \ref{p_order} \;
  }
  \While{not converged}{
    \While{element in $\hat{Q}$ and $\hat{S}$}{
    Evaluate calibrated skill element $q_{ij}$ using equation \ref{c_e};\;
    Replace the element in $\hat{Q}$ and $\hat{S}$ with a calibrated element \;
  }
  Update parameters learning rate $\alpha$ and hyper-parameter $\lambda$ \;
  }
  \caption{Knowledge Relation Importance Rank Calibration Method.}
\end{algorithm}

\paragraph{Exercise relation modeling.}
The exercise matrix modeling is designed based on two processes. Firstly, skill-exercise embedding is obtained by applying the GCN. Secondly, skill-exercise embedding, item difficulty, and the contingency table are incorporated to generate the exercise relation matrix.

For the GCN model, high-order neighbor information can be encoded by combing the skill relation matrix and the Q-matrix with the heterogeneous interactions. The GCN model consists of numerous convolutional layers, and each layer can be updated by the states of itself and the neighbors of nodes. The \emph{i}th node in the graph donated as $node_i$ indicating the skill state s$_i$ or exercise state e$_i$. The neighboring nodes of $node_i$ are denoted as a set of nodes: Node(i). As a result, the $\imath$ th layer of the graph convolutional network can be updated as follows:
\begin{equation}
    node^{\imath}_i = RELU(\sum_{j\in {i}\cup Node(i)}^{} w^{\imath}_{i}node^{\imath-1}_i + b^{\imath}_i)
\end{equation}
where w$^\imath$ and b$^\imath$ infer as the weighted matrix and bias of the GCN layer and the RELU() indicates the activation function accepted in the GCN model. Then the skill-exercise embedding: $\hat{e}$ is used for estimating the implicit relations among questions by calculating the inner product of questions:
\begin{equation}
    simi_{i,j} = \frac{\hat{e_i}\cdot  \hat{e_j}}{|\hat{e_i}||\hat{e_j}|}
\end{equation}
The exercises' similarity further incorporates the item difficulty with the previous students' performance to generate $R^E$. 
For the item difficulty modeling, the students' incorrect interactions can intuitively represent the item difficulty of the exercises involved in the student interactions. And the students repeat the same questions by utilizing their skills in different timestamps. This behavior also can demonstrate the cognitive difficulty of these exercises. In order to model this situation, the cognitive question difficulty for a student: s can be defined as follows:
\begin{align}
\begin{split}
\Psi _{s,q,t}= \left \{
\begin{array}{ll} 
     \left [ \frac{|\{R_s==0\}|_{0:t}}{|Q|_{0:t}}*4  \right ] &  if|N_v|_{0:t}\ge 5  \\          
     5                                 & otherwise
\end{array}
\right.
\end{split}
\end{align}
where $\Psi _{q,t}$ indicates each student's cognitive difficulty of the question set at timestamp t. The cognitive difficulty is divided into 5 levels including very hard, hard, medium difficulty, relatively easy, and easy. The number ranging from 0 to 5 is accepted to indicate the corresponding levels. The $|Q|$ denote the set of questions before timestamp t and $R_s$ refers to the student's response to the same questions. A zero in the $R_s$ indicates the student provides a wrong answer for a question. If a learner attempt to answer a question fewer than five times, the cognitive difficulty of this question is directly quantified into 5. Then according to the different learners ' cognitive difficulty of questions, the average cognitive difficulty for different learners on the same question is defined as the item difficulty: $\varphi(q)$ after processing the cognitive difficulty for different learners into the GCN.
\begin{equation}
    \varphi(q) = \frac{1}{|\Psi _{s,q,t}|}\sum_{\Psi(s_i,q,t)\in \Psi(s,q,t)  }^{} \Psi(s_i,q,t) 
\end{equation}
Then, according to the item difficulty: $\varphi_q$, the similarity of question difficulty can be modeled as follows:
\begin{equation}
    diff(q_i,q_j) = \frac{1}{1+\varphi(q_i)-\varphi(q_j)}
\end{equation}
In order to incorporate the previous interaction, the students' performance on question pair $q_i$ and $q_j$ is summarized in the contingency table. The students' correct and incorrect responses are interpreted as the mastery indicators of the questions referring to Table~\ref{contingency}. When a question pair appears more than once in the previous student interactions, the latest occurrence is taken into consideration. According to the contingency table, seven evaluation metrics, measuring the association between two variables, are developed to measure the relationship between the question pair: $e_i$ and $e_j$ referring to Table ~\ref{coff}. A threshold is imposed to control the sparsity of relations of exercises. The exercise relation matrix based on the contingency table is denoted as $W^R$, $R \in \{SK, Kappa, Kappa^{'}, Phi, Yule, Ochiai, Sokal, Jaccard.\}$.  
\begin{align}
\begin{split}
  \left \{
\begin{array}{ll} 
     W^R_{i,j} = max(W^R_{i,j},W^R_{j,i}), W^R_{j,i}=0 &  W^R_{i,j}\ge W^R_{j,i}  \\          
     W^R_{j,i} = max(W^R_{j,i},W^R_{i,j}), W^R_{i,j}=0                                 & otherwise
\end{array}
\right.
\end{split}
\end{align}
\begin{table}[ht] 
\centering
\begin{tabular}{c|c|c|c|c}
\toprule
\multicolumn{2}{c}{\multirow{2}{*}{}} & \multicolumn{3}{c}{\textbf{exercise i}}
\\
\midrule
& & F & T & total\\ 
\multirow{2}{*}{\textbf{exercise j}}& F & a & b & a+b\\
\multirow{2}{*}{}& T& c & d & c+d\\ 
&total& a+c & b+d & a+b+c+d\\
\bottomrule
\end{tabular}
\caption{The contingency table for exercise \emph{i} and exercise \emph{j}. The labels: "F" and "T" present the student answering the exercise incorrectly or correctly. \label{contingency}}
\end{table}

\begin{table}[ht]
\centering
\begin{tabular}{|l|l|}
\hline
Cohen's Kappa & $W_{e_i, e_j}^{Kappa}$ = 2(ad-bc)/{(a+b)(b+d)+(a+c)(c+d)} \\

Adjusted Kappa & $W_{e_i, e_j}^{Kappa^{'}}$ = 2(ad-bc)/{(a+c)(c+d)} \\

Phi coefficient &  $W_{e_i, e_j}^{Phi}$ = (ad-bc)/$\sqrt{(a+b)(b+d)(a+c)(c+d)}$ \\

Yule coefficient & $W_{e_i, e_j}^{Yule}$ = (ad-bc)/(ad+bc)\\

Ochiai coefficient & $W_{e_i, e_j}^{Ochiai}$ = a/$\sqrt{(a+b)(a+c)}$\\

Sokal coefficient &  $W_{e_i, e_j}^{Sokal}$ = (a+d)/$\sqrt{(a+b+c+d)}$\\

Jaccard coefficient &  $W_{e_i, e_j}^{Jaccard}$ = a/(a+b+c)\\
\hline
\end{tabular}
\caption{\label{coff}Seven evaluation metrics. These metrics are designed to explore the association between two variables.}
\end{table}
Finally, the relation of exercise: i with exercise: j is calculated as follows: 
\begin{align}
\begin{split}
  A_{i,j} = \left \{
\begin{array}{ll} 
      \mu_1 simi_{i,j} + \mu_2 diff(q_i,q_j) + \mu_3 W^R_{i,j}  & if \quad \mu_1 simi_{i,j} + \mu_2 diff(q_i,q_j) + \mu_3 W^R_{i,j} \ge \Theta   \\          
      0                              & otherwise
\end{array}
\right.
\end{split}
\end{align}
where $\Theta$ is a threshold to control the sparsity of the exercise relation matrix. Then Given the past exercises: (${e_1,e_2,...e_{n-1}}$) and the next exercise: $e_n$, the exercise relation matrix is defined as $R^E$ = [$A_{e_n,e_1}$, $A_{e_n,e_2}$,...$A_{e_n,e_{n-1}}$]. Finally, the exercise relation matrix is applied as the input of the Position-Relation-Forgetting attention mechanism.

\paragraph{Position-Relation-Forgetting attention mechanism.}
The Position-Relation-Forgetting attention mechanism includes the relative position attention layer, the relation attention layer, and the forgetting layer. The relative position attention accepts the relative distance between input elements. $x_i$ and $x_j = (x_1,x_2..x_{n-1})$ are served as the model inputs to track the student state. And edge vectors between $x_i$ and $x_j$ are presented as $a_{i,j}^v, a_{i,j}^K$ to extract relative position representations. The edge vectors are clipped to a maximum absolute value of k: clip(x,k) = max(-k, min(k,x)). And corresponding relative position representations are $W^K = (W^k_{-k}....W^K_{k})$ and $W^V = (W^V_{-k}....W^V_{k})$ respectively. The outputs of the relative position attention mechanism are new sequences $Z$. The process can refer to the following equations:
\begin{equation}
a^k_{i,j}  =  W^k_{clip}(j-i,k) \quad a^V_{i,j}  =  W^V_{clip}(j-i,k)  
\end{equation}
\begin{equation}
   a_{i,j} = \frac{exp(e_{i,j})}{\sum_{i=1}^{n}exp(e_{i,k}) } \quad e_{i,j} = \frac{x_iW^Q(x_jW^k)^T + x_iW^Q(a_{i,j}^K)^T}{\sqrt{d_z}}
\end{equation}
\begin{equation}
    Z_i = \sum_{j=1}^{n} a_{ij}(x_jW^V) \label{pos_output}
\end{equation}
where W$^Q$, W$^K$, and W$^V$ are the query, key, and value matrices respectively and d$_z$ is the dimension of the new sequence of Z. Then, the relation attention predicts student performance on the next interaction by combining the outputs of the relative position attention mechanism. 

The relation attention layer incorporates the output of the formula (\ref{pos_output}) with the exercise relation matrix to predict student performance on the next interaction. 
\begin{equation}
    e_i = \frac{E_{e_n}W^Q(Z_jW^K)^T}{\sqrt{d} } \quad \alpha_i = \frac{exp(e_i)}{\sum_{k=0}^{n-1} exp(e_k) }
\end{equation}
\begin{equation}
    \gamma_i = \delta  \alpha_i +(1-\delta  )R^E_i \quad H = \sum_{i=1}^{n-1}\gamma_i{Z}_iW^v  
\end{equation}
where W$^Q$, W$^K$, and W$^V$ represent the query, key, and value matrices of the attention mechanism.
Then applying the output of the relation attention layer is treated as the input of the forgetting layer based on learning theory in the educational field. The relative time intervals between past and next interactions are compared as $\bigtriangleup_i = t_{n} - t_i$. The final outputs of the Position-Relation-Forgetting attention mechanism incorporating forgetting behavior,$R^F$, is computed as follows:
\begin{equation}
    R^F = [ \xi_1  e^{-\xi_2  \bigtriangleup_1}, \xi_1 e^{- \xi_1 \bigtriangleup_2} ...\xi_1 e^{- \xi_2 \bigtriangleup_n}] \quad  O = \delta_{F} H+(1-\delta_{F})R^F
\end{equation}
where  $\xi_1$ and $\xi_2$ are hyper-parameters.

\paragraph{Student performance prediction layer.}
The student performance prediction layer contains the pointwise Feed-Forward (FFN) and probability prediction layer. The FFN can be computed as follows referring to(~\ref{ffn}). $W_l$ and $W_s$, $b_l$ and $b_s$ are weighted matrices and bias vectors respectively.
\begin{equation}
    F= ReLU(OW_l+b_l)W_s+b_s \label{ffn}
\end{equation} 
The probability prediction layer predicts the probability of the student's performance by accepting function: $\sigma()$ on the basis of the FFN. P denotes as the probability that the students provide correct answers in the next interaction. The W and b are trainable parameters.
\begin{equation}
 p = \sigma (FW + b)
\end{equation}

\section*{Implementation and experimental results\label{er}}
\paragraph{Implementation details.}
\emph{Framework setting}. The model dimension of attention, the max sequence length, and the training batch size are 200. The dropout rate of the NGFKT model is 1e-2. And the hyper-parameters, including the $\lambda$, $\Theta$, are 1 and 0.65 respectively. The parameters in the exercise relation modeling: $\mu_1$, $\mu_2$, and $\mu_3$ are 0.1, 0.2, and 0.7 respectively. The other parameters that are not specified involved in the process of the training model are normally initialized as 0. 
\paragraph{Evaluation methodology.}
\emph{Metrics}.
The prediction task is evaluated in a binary classification scenario, i.e.,  whether or not an exercise is performed correctly. As a result, the Area Under Curve (AUC) and Accuracy(ACC) are accepted to measure the prediction performance of students. AUC or ACC values of 0.5 usually indicate that the result was determined at random. The greater the knowledge tracing performance, the higher the value of AUC or ACC. The cross-entropy is accepted as the loss function of the NGFKT model. 

The Performance Stability metric(PS) is used to specifically compare the performance of the baseline models with the NGFKT model in the testing phase. The performance of the model: M is stable when the M can consistently outperform other models in most testing batches. Based on this idea, the PS is designed based on the performance rank.
For instance, if the NGFKT model outperforms the DKT model and DKT+ model in 96 testing batches. However, the performance of the NGFKT model is worse than the DKT+ model in 4 testing batches. The performance rank of the NGFKT model is 1 in 94 testing batches and 2 in 4 testing batches. Then the PS of the NGFKT model is 97.32\% referring to the following formulation. The $N_{Batch}$ and $N_{model}$ are the number of the testing batches and models in this paper.
\begin{equation}
    PS(M) = \sum_{i=0}^{N_{Batch}} \frac{N_{model}-rank(M,i)+1}{N_{model}} 
\end{equation}

\paragraph{Datasets.} In order to demonstrate the performance of the NGFKT model in the small dataset and the large dataset, two types of public educational datasets were accepted to verify the performance of the NGFKT model in terms of AUC, ACC, and PS. And the NGFKT-PE and NGFKT-RM, the variants of the NGFKT model, present that the NGFKT model removes the relative position attention mechanism or relation modeling respectively. The statistical information of the datasets is provided in Table~\ref{q_statistic}. The first dataset is the large dataset: ASSISTments2012(ASSIST2012)[\href{https://sites.google.com/site/assistmentsdata/datasets/2012-13-school-data-with-affect}{https://sites.google.com/site/assistmentsdata/datasets/2012-13-school-data-with-affect}], which was gathered by the Assistemt Online Tutoring platforms. The ASSIST2012 contains 4193K records of 39K students. Each student answers an average of 107 questions. The second dataset is the small dataset: the Eedi dataset[\href{https://eedi.com/projects/neurips-education-challenge}{https://eedi.com/projects/neurips-education-challenge}], collected by the NeuralIPS platform, includes 233K records and 2064 students. And an average of 113 exercises are provided for each student. In this paper, tasks 3 and 4 of the NeuralIPS are applied to tracking the knowledge state of students.
\begin{table}[ht] 
\centering
\caption{The overview of ASSIST2012 and Eedi.\label{q_statistic}}
\begin{tabular}{c|c|c}
\toprule
\textbf{Statistic}	& \textbf{ASSIST2012} & \textbf{Eedi}\\
\midrule 
     Number of records & 4193631 & 233767 \\ 
     Number of students & 39364& 2064 \\ 
     Number of questions & 59761 & 948  \\ 
     Avg exercise record/student & 107& 113  \\ 
\bottomrule
\end{tabular}
\end{table}
\paragraph{Student performance prediction.} The experimental results are presented in Table ~\ref{q_res}. The AUC, ACC, and PS are computed to compare the performance of different models. In order to verify the prediction performance of student abilities, the training set and testing set are divided into 80\% records of the dataset and 20\% records of the dataset respectively. The NGFKT model is compared with the DKT model, the DKT+ model, the SAKT model, the GKT model, and two variants models of the NGFKT model: the NGFKT-PE and the NGFKT-RM. 

In two educational datasets, the NGFKT model is compared with the DKT model, the DKT+ model, the SAKT model, the GKT model, the NGFKT-PE model, and the NGFKT-RM model. The DKT+ model outperforms the DKT model due to the fact that two regularization terms are involved to solve the reconstruction problem and the consistency problem on both two datasets. The DKVMN model further improves the performance of the DKT+ model by incorporating the nonlinear representations on the ASSIST2012 and the Eedi. The GKT model considers the graph between the exercises and skills to improve the performance of the DKVMN model on the ASSIST2012. The SAKT model shows an improvement in results than the DKVMN model and the GKT model by considering the relation modeling on the ASSIST2012. However, the relationship between exercises and skills is ignored in the SAKT model. The NGFKT-RM and the NGFKT-PE consider the heterogeneous interactions between students, exercise, and skills to improve the performance of the SAKT model on the ASSIST2012 and the Eedi. The NGFKT-PE model possesses better performance than the NGFKT-RM indicating that relation modeling is a more important part than the relative position layer on the big dataset: ASSIST2012. The NGFKT-RM model outperforms the NGFKT-PE showing that adding a relative position attention layer is a more efficient method than the relation modeling to improve the performance on the small dataset. The NGFKT model performs consistently better than all baseline models by incorporating the relative distance representations and relation modeling to predict the performance.

The different performances of models on the two datasets can be traced to the fact that there is a huge gap in the number of interactions provided by students. Therefore, some models are suitable for handling the large dataset(ASSIST2012) such as the SAKT model and the NGFKT-PE model. Some models are good at handling the small dataset(Eedi) such as the DKT+ model and the NGFKT-RM model.
\begin{table}[ht]
\centering
\caption{Comparison of results of baseline models with the Neural Graph Forgetting Knowledge Tracing model(NGFKT). The NGFKT outperforms all baseline models in terms of AUC, ACC, and PS. \label{q_res}}
\begin{tabular}{ccccccc}
\toprule
\multicolumn{1}{c}{\multirow{2}{*}{}}&\multicolumn{3}{c}{\textbf{ASSIST2012}}&\multicolumn{3}{c}{\textbf{Eedi}}\\
\midrule
& AUC & ACC & PS & AUC & ACC & PS \\
    DKT& 0.712 & 0.679 &0.142&0.489 & 0.489 & 0.179  \\
    DKT+& 0.722 & 0.685 &0.232&0.584 & 0.566 & 0.501\\
    DKVMN& 0.701 & 0.686 &0.392&0.701& 0.640 & 0.829\\
    SAKT& 0.736 & 0.692 &0.732&0.495& 0.495 & 0.216\\
    GKT & 0.702 & 0.687  & 0.491& 0.566 & 0.542 & 0.354\\
    NGFKT-RM & 0.740 & 0.678  & 0.678 & 0.684 & 0.637 & 0.804\\
    NGFKT-PE & 0.753 & 0.689  & 0.860 & 0.691 & 0.641 & 0.679\\
     \textbf{NGFKT}& \textbf{0.776} & \textbf{0.704} &  \textbf{0.960} & \textbf{0.710} & \textbf{0.673} &\textbf{0.937}  \\ 
\bottomrule
\end{tabular}
\end{table}

\paragraph{Cold start problem.}The cold start issue also impacts knowledge tracing. The results are examined in two cold start scenarios, i.e., training with data from a small number of students and new students with abbreviated exercise sequences \cite{2020Cold}.
\begin{enumerate}
    \item The knowledge tracing model is trained using data from a limited number of students, and it is then applied to completely new, untested samples.
    \item When new students enroll in an online learning system, they often have short exercise sequences since there aren't enough exercise recordings to provide the knowledge tracing model with enough information about them.
\end{enumerate}

\begin{figure}[ht]
	\centering  
	\subfigbottomskip=2pt 
	\subfigcapskip=-5pt 

	\subfigure[AUC on the Eedi dataset is compared when the number of the students is 10\%, 12\%, 14\%, 16\%, 18\%, and 20\% respectively. The Neural Graph Forgetting Knowledge Tracing Model outperforms the other two models.\label{cold}]{
		\includegraphics[width=0.48\linewidth]{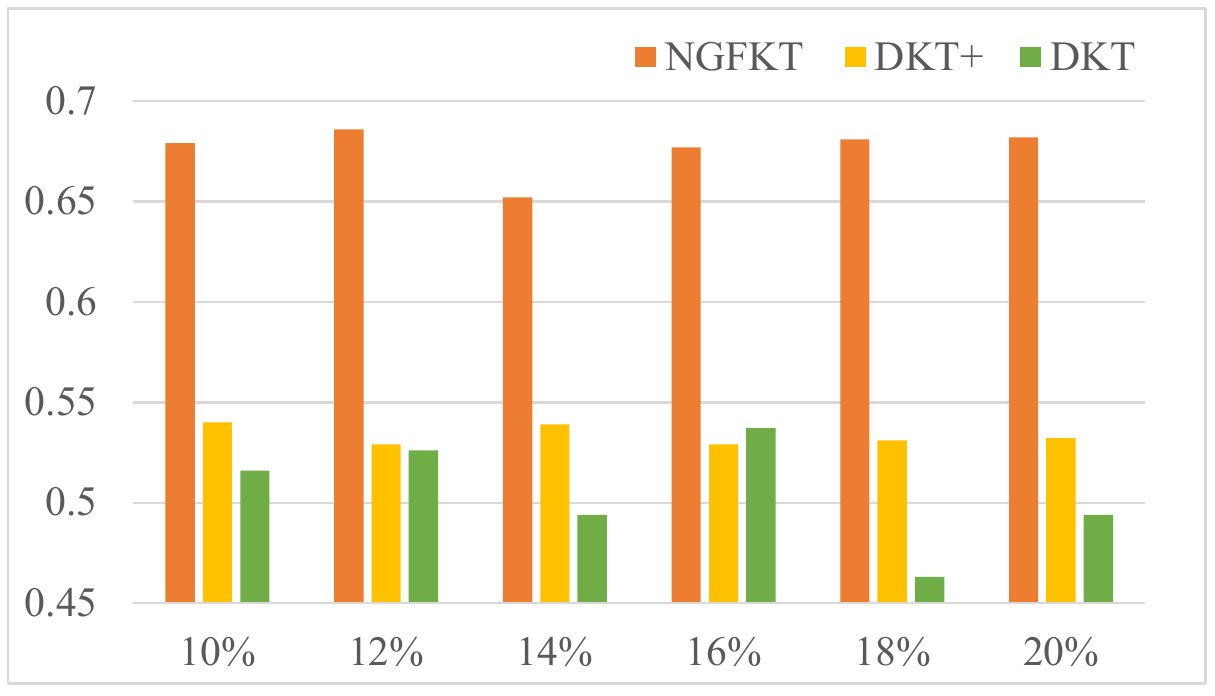}}
	\subfigure[AUC on the Eedi dataset is compared when students are provided with 50, 75, 100, 125, 150, 175 and 200 answer records respectively. The Neural Graph Forgetting Knowledge Tracing Model achieves better results than the other two models.\label{cold2}]{
		\includegraphics[width=0.48\linewidth]{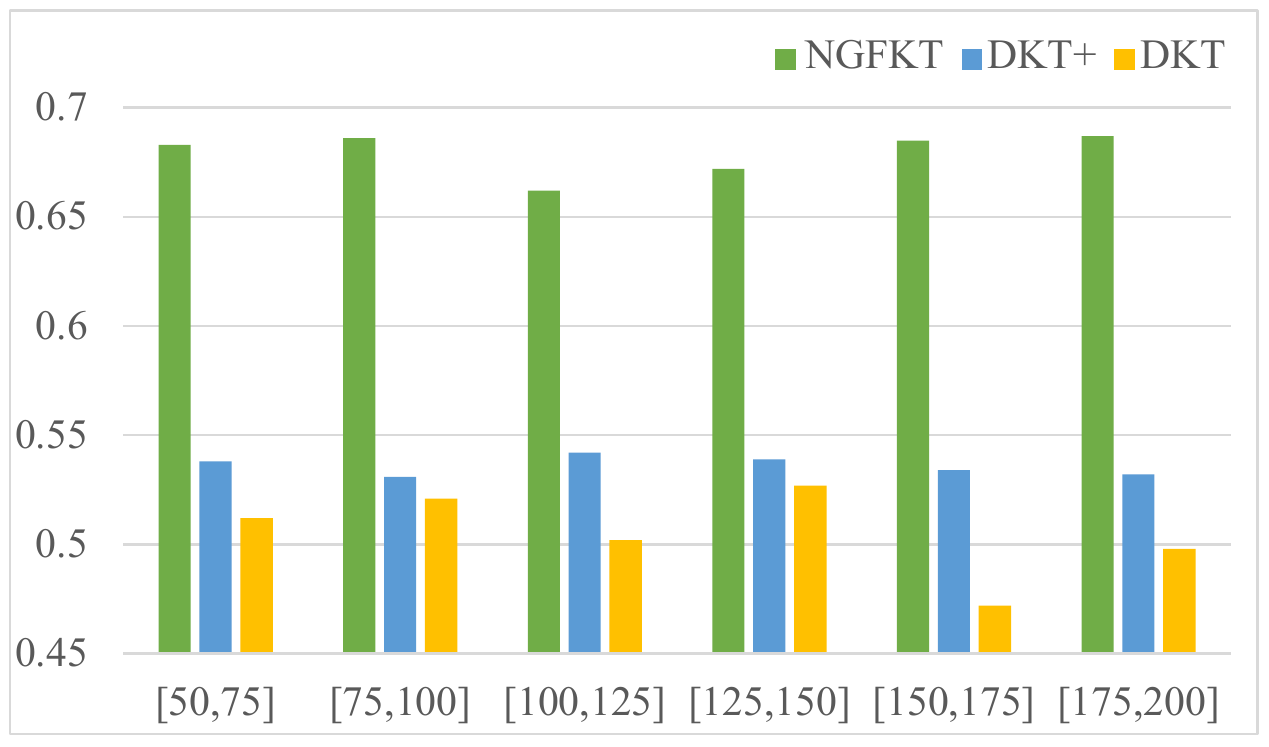}}
	  \\
\end{figure}

In scenario 1, student populations ranging from 10\% to 20\% of the training dataset are tested to evaluate the prediction performance of NGFKT, DKT, and DKT+ referencing Figure~\ref{cold}. The NGFKT model outperforms the DKT model and DKT+ model on the Eedi dataset and the performance of the NGFKT model increases when more records of students are implemented.

\begin{figure}[ht]
\centering
\includegraphics[width=0.8\linewidth]{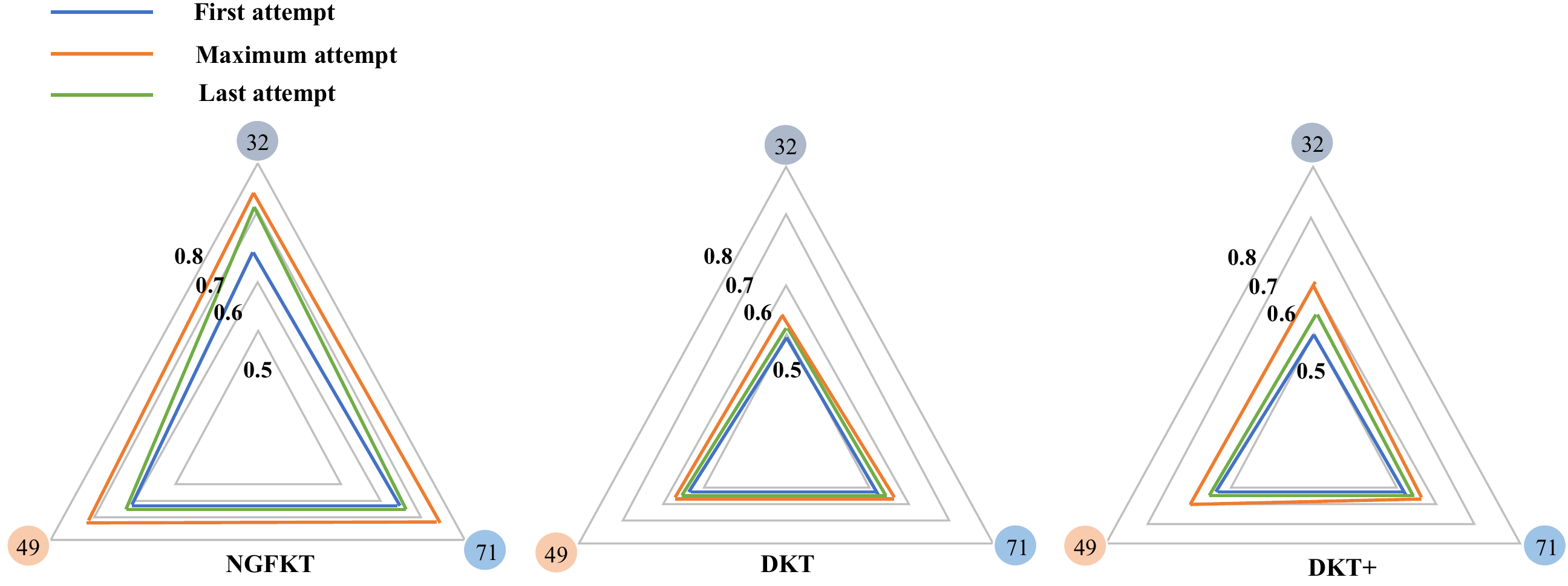}
\caption{The radar diagram. The NGFKT model outperforms the DKT model and the DKT+ model in tracking the student knowledge state. The "32", "49", and "71" are three skill ids and are presented with three different colors. The average prediction accuracy of the NGFKT model is around 70.5\%.}
\label{radar}
\end{figure}

In scenario 2, which contains new students who have short exercise sequences, the training data are separated into six groups. Each group has a distinct range of exercise sequences, such as (50, 75], (75, 100], (100, 125], (125, 150], (150, 175], and (175, 200]. The lengths of the exercises from the original exercise sequence are sampled to generate each exercise sequence in the training data. Given that students in the first group have the fewest exercise answering records, it is clear that this situation is the most challenging for the student. In Figure~\ref{cold2}, the effectiveness of the various techniques is compared. On the Eedi datasets, the NGFKT model shows better performance in this scenario than DKT and DKT+.
\paragraph{Knowledge state prediction visualization.} 
Knowledge state prediction visualization is regarded as an important application of knowledge tracing models for online educational systems. We will show that our proposed model: the NGFKT model can capture the student knowledge state correctly compared with two standard knowledge tracing models: the DKT model and the DKT+ model. Specifically, Figure~\ref{radar} indicates the knowledge state traced by the NGFKT model of the same student. The general knowledge evolving process of the knowledge state is consistent with the student learning process. When the student first attempts the exercise, the knowledge state reaches the minimum level. The student continues to learn skills: "32", "49", and "71" and continuously deepens his proficiency in knowledge points. Finally, the student knowledge state achieves the maximum, which is shown by the increased areas of the radar diagram. During the latest attempt of the student, the knowledge proficiency of the student presents some reduction considering the student forgetting behavior. But, knowledge proficiency is still improved by continuously practicing the skills compared with the first interaction with the student.

Referring to Figure~\ref{radar}, the NGFKT model also outperforms the DKT model and the DKT+ model because the NGFKT model further incorporates the relation modeling that is generated by the GCN model. The DKT+ model achieves better results than the DKT model, which indicates that adding two regularization terms can further improve the performance of tracing the knowledge proficiency of the student.
\section*{Conclusions and future work}
In this paper, a novel knowledge tracing model: NGFKT is proposed to track the student knowledge state by incorporating the relation modeling, skill relation matrix, Q-matrix, and relative distance representations. The NGFKT model can track the student knowledge state accurately even with small amounts of interactions. Specifically, this paper applies the KRIRC method to calibrate the skill relation matrix and Q-matrix and these two matrices served as the input of the GCN to generate the exercise embedding and skill embedding. The skill-exercise embedding, the item difficulty, and the contingency table are incorporated to generate the final exercise relation matrix. Finally, utilizing the Position-Relation-Forgetting Attention layer outputs the predicted results. The experiments conducted on two public datasets indicate that the NGFKT model can track the student knowledge state efficiently. A combination of explainable and predictive 
power in the NGFKT model will contribute to the better design of the Online Educational System. In the future, we plan to take the exercise texts into the design of the knowledge tracing model and consider more student behaviors, such as the guessing factor, in relation modeling.
\section*{Data availability}
The datasets involved in this paper and the proposed methods are available in the NGFKT repository.

[\href{https://github.com/Destiny123456qwer/NGFKT}{https://github.com/Destiny123456qwer/NGFKT}]

\section*{Funding}
This research work presented in this paper was supported by the National Natural Science Foundation of China (Nos. 62177022 and 61901165), the AI and Faculty Empowerment Pilot Project (No. CCNUAI\&FE2022-03-01), the Collaborative Innovation Center for Informatization and Balanced Development of K-12 Education by the MOE and Hubei Province (No. xtzd2021-005), Natural Science Foundation of Hubei Province (No. 2022CFA007), and the National Natural Science Foundation of China (No. 61501199).

\section*{Competing interests}
The authors declare no competing interests.

\bibliography{sample}
\end{document}